\documentclass[sn-basic]{sn-jnl}

\usepackage{chngcntr}
\usepackage{graphicx}%
\usepackage{tabularx}
\usepackage{multirow}%
\usepackage{amsmath}%
\usepackage{amssymb}
\usepackage{amsfonts}
\usepackage{amsthm}%
\usepackage{mathrsfs}%
\usepackage[title]{appendix}%
\usepackage{xcolor}%
\usepackage{textcomp}%
\usepackage{manyfoot}%
\usepackage{booktabs}%
\usepackage{algorithm}%
\usepackage{algorithmicx}%
\usepackage{algpseudocode}%
\usepackage{listings}%
\usepackage{subcaption}

\theoremstyle{thmstyleone}%
\theoremstyle{thmstyletwo}%

\theoremstyle{thmstylethree}%

\newcolumntype{Y}{>{\centering\arraybackslash}X}

\begin{document}

\title[Article Title]{Speech Emotion Recognition Leveraging OpenAI's Whisper Representations and Attentive Pooling Methods}


\author[1]{\fnm{Ali} \sur{Shendabadi}}\email{alishendabadi@ut.ac.ir}

\author[1]{\fnm{Parnia} \sur{Izadirad}}\email{parniaizadirad@ut.ac.ir}

\author*[1]{\fnm{Mostafa} \sur{Salehi}}\email{mostafa\_salehi@ut.ac.ir}

\author[2]{\fnm{Mahmoud} \sur{Bijankhan}}\email{mbjkhan@ut.ac.ir}

\affil[1]{\orgdiv{Faculty of Intelligent Systems Engineering}, \orgname{University of Tehran}, \orgaddress{\street{N.Kargar}, \city{Tehran}, \country{Iran}}}

\affil[2]{\orgdiv{Faculty of Literature and Humanities}, \orgname{University of Tehran}, \orgaddress{\street{Enghelab}, \city{Tehran}, \country{Iran}}}




\abstract{Speech Emotion Recognition (SER) research has faced limitations due to the lack of standard and sufficiently large datasets. Recent studies have leveraged pre-trained models to extract features for downstream tasks such as SER. This work explores the capabilities of Whisper, a pre-trained ASR system, in speech emotion recognition by proposing two attention-based pooling methods, Multi-head Attentive Average Pooling and QKV Pooling, designed to efficiently reduce the dimensionality of Whisper representations while preserving emotional features. We experiment on English and Persian, using the IEMOCAP and ShEMO datasets respectively, with Whisper Tiny and Small. Our multi-head QKV architecture achieves state-of-the-art results on the ShEMO dataset, with a 2.47\% improvement in unweighted accuracy. We further compare the performance of different Whisper encoder layers and find that intermediate layers often perform better for SER on the Persian dataset, providing a lightweight and efficient alternative to much larger models such as HuBERT X-Large. Our findings highlight the potential of Whisper as a representation extractor for SER and demonstrate the effectiveness of attention-based pooling for dimension reduction.
}




\maketitle

\section{Introduction}\label{secIntro}

In recent years, human-computer interaction has entered a new phase. AI assistants such as ChatGPT have become an everyday tool for humans, raising the need for machines to better understand humans and their needs. A significant factor in achieving this is the capability of these systems to correctly detect human emotions \cite{robots}. In spoken interactions, emotions are conveyed through the tone of the utterance; more specifically, the manner in which the words are spoken and their semantic meaning. The task of Speech Emotion Recognition (SER) is to detect the emotions in an utterance using features extracted from the speech signal. An SER system that better comprehends human emotions enables more appropriate responses to users and helps anticipate their needs.

\begin{figure*}[h!]
  \centering
  \includegraphics[width=\textwidth]{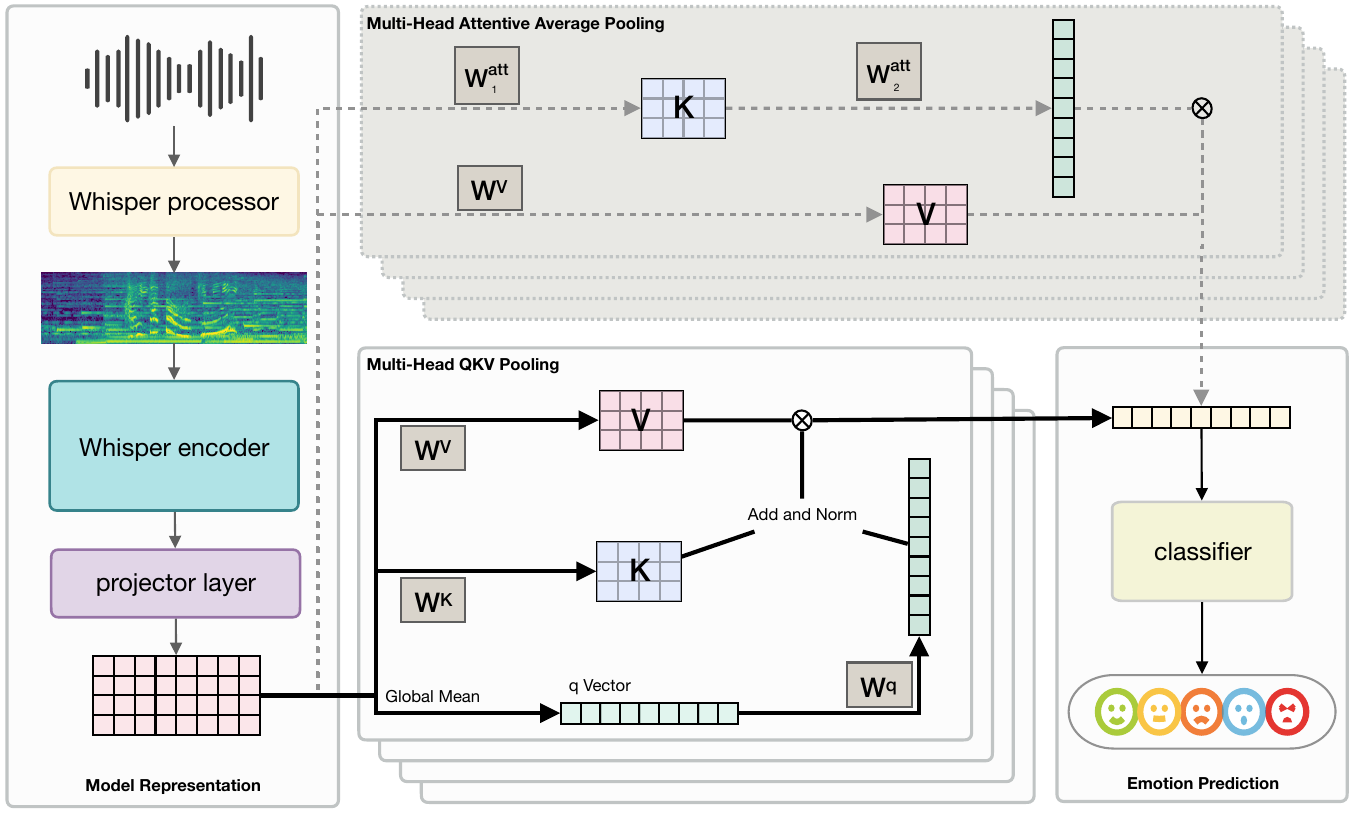}
  \caption{
  Demonstration of the Multi-head Attentive Average Pooling and Multi-head QKV Pooling pipeline for SER. After extracting speech representations using Whisper encoders, a multi-head pooling method is applied to reduce the dimensionality of the representation matrix. The stacked rectangles represent attention heads. The outputs from all attention heads are concatenated and subsequently projected through a weight matrix ${W^o}$, producing a final 256-dimensional vector that serves as the input to the classifier.
  }
\end{figure*}

In recent studies, pre-trained models have been adapted for extracting features as opposed to traditional features such as MFCC. These studies have achieved state-of-the-art results by fine-tuning pre-trained models on an SER corpus. Some other studies treat the pre-trained model as an upstream model for feature extraction by freezing all layers and building a downstream model for predicting emotions based on the representations produced by the pre-trained model.

In this study, we use Whisper \cite{whisper}, a pre-trained ASR model, for feature extraction by freezing the encoder layers. We then apply a Dimension Reduction Module (DRM) for pooling a single vector from the representation matrix. We propose two attention-based pooling methods: Attentive Pooling and QKV Pooling. In Attentive Pooling, we initially follow the experiment carried out in \cite{nasersharif}, where each frame is given a weight to assess its importance for predicting the emotion in an utterance. We further extend this approach by applying it in a multi-head manner. In QKV Pooling, the query is conditioned on the global average pooled from the initial representation of the pre-trained model. Our results indicate that an attention-based pooling method can enable a smaller model to compete with the results of larger models.

The main contributions of this research are as follows:

\begin{itemize}
\item Proposing QKV Pooling, an attention-based pooling method that extracts a single vector from a high-dimensional representation matrix while minimizing information loss. Additionally, we adapt Attentive Pooling on Whisper. Our results show that Whisper Small with QKV achieves performance comparable to Whisper Large V3 on both IEMOCAP and ShEMO, demonstrating a consistent and promising trade-off between cost and accuracy.

\item Exploring the effectiveness of the Whisper encoder as a representation extractor for SER in Persian, a lower-resource language.

\item Analyzing the different layers of the Whisper encoder and comparing their ability to create effective speech representations.
\end{itemize}

Although other models achieve higher accuracy, as discussed in Section \ref{sec: result and exp}, our study emphasizes that an effective pooling method can enable a smaller pre-trained model to achieve results comparable to models that are hundreds of times larger. We release all the codes at \url{https://github.com/alishendabadi/AttentivePoolingWhisperSER}. 

\section{Related Works and Background}\label{secRelated}

Early SER systems relied on traditional machine learning methods such as Hidden Markov Models (HMMs), Gaussian Mixture Models (GMMs), Support Vector Machines (SVMs), and Decision Trees \cite{pepino}. In these methods, researchers employed hand-crafted acoustic features like MFCCs, Linear Predictive Cepstral Coefficients (LPCCs), energy, pitch, and speaking rate. While capable, these approaches faced fundamental limitations in capturing the complexity and individuality of human emotion \cite{dstm}.

More recent studies have explored deep learning approaches and achieved good experimental results. Recurrent Neural Networks (RNNs) and Long Short-Term Memory (LSTM) networks have been used to learn high-level feature representations by processing speech signals sequentially \cite{guptalstm}. In DialogueRNN \cite{dialoguernn}, AGHMN \cite{aghmn}, and HiGRU \cite{higru}, researchers used Gated Recurrent Units (GRUs) to capture temporal dependencies and contextual relationships in emotional speech. In other studies \cite{cnnrnn1, cnnrnn2, cnnrnn3}, Convolutional Neural Networks (CNNs) and RNNs were combined to model and extract frequency and temporal features from emotional speech, respectively.

Despite achieving significant results, these models still faced several challenges. CNNs suffered from an inability to capture long-term dependencies due to their fixed window sizes. RNNs, on the other hand, were prone to issues such as gradient vanishing or exploding problems when processing lengthy sequences, which limited their overall understanding of emotions. Additionally, combining CNN and RNN structures often increased model complexity. This could lead to higher computational costs, more time-consuming training, and an increased risk of overfitting, especially on small datasets. 

In response to the above issues researchers have used Transformer models for modeling SER task. These self-attention based models are better equipped to process lengthy input sequences and capture their global dependencies.


An ongoing problem throughout SER research has been the lack of standard datasets. The available datasets are usually small and unbalanced. For this reason, researchers have adopted transfer learning in their work, where pre-trained models are used to learn meaningful representations applicable to downstream tasks. Pre-trained models like Wav2Vec, HuBERT, and Whisper have significantly impacted various downstream speech processing tasks, including SER, by providing rich, contextualized speech representations. For example, Jiao et al. \cite{mfhca} used HuBERT features and log Mel-spectrograms in MFHCA, an SER method based on a Multi-Spatial Fusion module (MF) and a Hierarchical Cooperative Attention module (HCA), to identify emotion-related regions and integrate features. HuBERT was also used for feature representation in DSTM \cite{dstm}, a Transformer-based network with dynamic-static feature fusion, using a locally adaptive multi-head attention module (DTM) for dynamic local features and a global static attention module (STM) for global features.

Although MFHCA and DSTM have achieved benchmark results using Hubert x-large representations, the large number of parameters in this model remains a challenge in handling high time and space complexity utilizing it. More recent pre-trained models are more robust in speech recognition. With these models researchers were able to lower the computation cost without sacrificing accuracy. For example Qu et al. \cite{breaking} developed SER-Whisper, a framework that integrates a frozen Whisper encoder with specialized transformer-based classification layers, achieving high accuracy on the RAVDESS \cite{ravdess_dataset} dataset. Similarly, Sankaran et al. \cite{hatespeech} and Ma et al \cite{emobox} compared Whisper-based representations with other pre-trained models such as Wav2vec 2.0, WavLM \cite{wavlm_model} and data2vec \cite{data2vec_model} showing that Whisper consistently outperforms alternatives in both audio abuse detection and SER tasks.

Further efforts have focused on adapting Whisper for emotion-related applications. Chou et al. \cite{tiny_whisper_ser} proposed Whisper-SER, which leverages Whisper Tiny with a lightweight adapter for multi-label SER. Their approach employed a two-stage training strategy with weighted losses to mitigate conflicts between ASR and SER objectives. However, they noted that Whisper Tiny exhibited higher word error rates (WER) on emotional speech datasets like MSP-PODCAST compared to neutral utterances. Feng et al. \cite{foundation} extended Whisper's role to SER data curation by generating transcriptions and utilizing encoder outputs as representations. They also explored the use of large language models (LLMs), including LLaMa 2, Falcon, and Flan-T5 XXL, for emotion annotation and data augmentation. Although LLMs struggled in zero-shot emotion annotation, combining multiple models and incorporating limited human feedback substantially improved annotation quality and SER performance.

To efficiently process the high-dimensional outputs of pre-trained speech models, dimension reduction has emerged as a critical technique for mitigating computational costs and overfitting risks. Nasersharif et al. \cite{nasersharif} a DRM, which integrates attentive average pooling, maxout activation, and linear layers to extract relevant features while reducing dimensionality. This approach enables dynamic selection of informative blocks and feature fusion via attention mechanisms, thereby enhancing SER performance.

A parallel body of work has examined Wav2vec 2.0 for SER, with particular focus on leveraging intermediate transformer block outputs. Studies have shown that early blocks encode acoustic features, while later ones capture linguistic information. Pepino et al. \cite{pepino} proposed combining multiple Wav2vec 2.0 layers through trainable weights, achieving superior SER performance and providing insights into the relative contributions of different layers. Nasersharif et al. \cite{nasersharif} worked on IEMOCAP and ShEMO \cite{ShEMO_dataset} and demonstrated that intermediate layers outperform both initial and final ones for emotion recognition.

\section{Methodology}\label{secMethod}
This section describes the three main elements in our study: 1. A pre-trained model that creates a representation from an audio sample. 2. A pooling method based on an attention mechanism that limits the loss of important information during dimension reduction. 3. A classifier which predicts the most likely emotion.

\subsection{Model Representation}\label{secWhisper}
In this study, we use Whisper features for representing audio samples. Whisper has shown state-of-the-art performance in many speech-related tasks, including SER. We also selected this model because of its multilingual capabilities, as we experiment on both English and Persian. Here, we give a brief review of Whisper and then explain the feature extraction process in detail.

\subsubsection{Whisper}
Whisper is a weakly supervised multilingual and multitask speech recognition model pre-trained on 680,000 hours of labeled audio data. Whisper uses an encoder-decoder Transformer architecture. First, audio inputs are re-sampled to 16,000 Hz and an 80-channel log-Mel spectrogram of the audio segments is computed and globally normalized. These spectrograms are then passed through two convolution layers. The encoder Transformer blocks then process the output of those convolutional layers using sinusoidal positional embeddings. Finally, the decoder part converts the representations to a sequence of tokens using learned positional embeddings and tied input-output representations.

What differentiates Whisper from other ASR models is its multilingual capabilities. Whisper is trained on a massive amount of data, including 117,000 hours covering 96 languages. Additionally, the diversity of speech data used for pre-training makes Whisper a robust model for low resource languages.

Whisper consists of three main parts; 1. two convolution layers which extract high level representation of mel-spectrograms created from raw audio by the Whisper processor. The mel-spectrograms are computed over 25 ms windows with a stride of 10 ms. 2. Encoder transformer blocks which process the spectrogram representations after adding positional embeddings. The encoder blocks use self-attention to create a sequence of encoder-hidden states. 3. Decoder transformer blocks that predict the next token using learned positional embeddings. The predictions are conditioned on the encoder-hidden states using cross-attention from encoder blocks in addition to previously predicted tokens using self-attention. The encoder and decoder blocks have the same width and number of layers.

\subsubsection{Feature Extraction}\label{subsec:overview}

For feature extraction, we use the encoder part of Whisper without updating its parameters. Utterances are first passed through the Whisper processor, which truncates each sample to 30 seconds and creates a log Mel-spectrogram from the raw audio. The Mel-spectrogram is processed by two convolutional layers using the GELU activation function. After the convolution layers, sinusoidal positional embeddings are added to the output. These embeddings can provide the model with information about the position of the audio features in the sequence. 

\begin{equation}
X = \text{Processor}(\mathbf{x}(t)), \quad X \in \mathbb{R}^{80 \times 3000}
\end{equation}

Finally, these enhanced representations are passed through the encoder Transformer blocks to process the final output. The final representation is a matrix of 1500 vectors with a dimension based on the model size. To achieve unified dimensionality for all whisper sizes a fully connected projector layer is applied. This layer maps the encoder output to a fixed model dimension and then normalizes each representation. In our experiments we set $d_{model}$ to 256 thus in all scenarios a representation matrix $\mathbb{R}^{1500 \times 256}$ will be obtained for classification. 

\begin{equation}
\label{eq2}
H = \text{Projector}(\text{WhisperEncoder}(X)),\qquad H \in \mathbb{R}^{1500 \times 256}
\end{equation}

\subsection{Attentive Pooling} \label{subsec:drm}

To make the Whisper representation ready for classification, we need to convert it into a one-dimensional vector, which causes some information loss. We propose two methods for this transformation: Multi-head Attentive Average Pooling and Multi-head QKV Pooling. In our experiments, QKV pooling generally outperforms Attentive Average Pooling, except for Weighted Accuracy on IEMOCAP. Both methods are included in this section to maintain coherence and organization of the paper.

\begin{equation}
a = \text{Pooling}(H), \quad a \in \mathbb{R}^{256}
\end{equation}

In equation \ref{eq2}, ${H}$ is the final model representation that was generated for each sample, and ${a}$ is the final vector that goes through the classifier after pooling.

\subsubsection{Multi-head Attentive Average Pooling}\label{section:MHSSAP}

In average pooling, all frames of an utterance have equal weight and therefore equal effect on the emotion prediction. However, in many cases, certain frames within an utterance contain more informative features for predicting emotion than others. To give appropriate weight to each frame, we adapt attentive statistics pooling, an attention-based pooling approach first introduced in \cite{attentive_statistics}. This mechanism is incorporated to assign importance to different frames within an utterance by integrating statistics pooling and the attention mechanism, creating a more robust speaker embedding.

Previous works \cite{nasersharif, attentive_statistics, nazari_brain} have achieved significant results by applying attentive statistics pooling for classification. In this study, we extend this approach by integrating a multi-head attention mechanism to improve the scaling accuracy. Specifically, each attention head is associated with a separate small neural network trained to compute the weights for each frame. 

As in \cite{attentive_statistics} and \cite{nasersharif}, we describe our attention network as mapping a value vector to an output, multiplying all of its elements by an attention weight.

\begin{equation}
e_t = f(\mathbf{h}_t W_1^{attn} + \mathbf{b})\mathbf{w}_2^{attn} + k
\end{equation}

, in which $W_1 \in \mathbb{R}^{d_{model}\times d_{hidden}}$ and $\mathbf{w}2 \in \mathbb{R}^{d{hidden}}$. $f(.)$ is the hyperbolic tangent activation function, and its output is passed through a dropout layer to prevent overfitting. Attention weights are calculated by being passed through a softmax function. Instead of performing a single attention function with $d_{hidden}$-dimensional value vectors and attention weights, it is computed on a set of queries packed together as a $V$ matrix and attention weights packed together in the $\mathbf{e}$ vector.

\begin{equation}
\text{Average Pooling}(\mathbf{e}, V) = \text{Softmax}\left({\mathbf{e}^T}\right) V
\end{equation}

The $V$ matrix is also formed by mapping each vector in the representation matrix to a vector of size $d_{hidden}$. All these vectors are grouped to form a $V$ matrix of size $\mathbb{R}^{1500 \times d_{hidden}}$.

\begin{equation}
V = HW^V, \quad W^V \in \mathbb{R}^{d_{model} \times d_{hidden}}
\end{equation}

The process described above is done 4 times in a 4-head manner using different sets of $W^V$. Output vectors from each head are concatenated together and mapped back to $d_{model}$ to form the pooled vector $a$.

\begin{subequations}
\begin{align}
&\text{Head}_i = \text{Average Pooling}(\mathbf{e}_i, V_i)\qquad \text{Concat}(\text{Head}_1, \dots, \text{Head}_n) W^{out}\\
&\mathbf{a} = \text{MultiHead}(\text{Head}_1, \dots, \text{Head}_n),\qquad \mathbf{a} \in \mathbb{R}^{d_{model}}
\end{align}
\label{formula: multiheadPooling}
\end{subequations}

,where $W^{out} \in \mathbb{R}^{(\#heads.d_{hidden}) \times d_{model}}$ and maps the dimension of the pooled vector to $d_{model}$

\subsubsection{Multi-head QKV Attention Pooling}\label{subsec: QKV}

In this method, we integrate a multi-head QKV attention mechanism to our pooling approach to reduce information loss during dimension reduction. In the QKV attention mechanism proposed by Vaswani et al. \cite{vaswani_attention}, the scaled dot product of $Q$ and $K$ is used to generate attention weights for each vector, with $Q$ derived from previously generated tokens. In our classification task, however, no prior outputs are available. To address this, we follow the pooling approach used in CLIP \cite{clip}, conditioning the query on the global average-pooled representation of the audio sample.

\begin{subequations}
\begin{align}
&\text{QKV Attention}(\mathbf{q}, K, V) =\qquad \text{Softmax}\left(\frac{\mathbf{q}K^T}{\sqrt{d_k}}\right) V \\
&\mathbf{\mu}(H) = \frac{1}{1500}\sum_{i=1}^{1500}\mathbf{h}_i, \quad \mathbf{\mu} \in \mathbb{R}^{256} \\
&\text{Head}_i = \text{QKV Att}(\mathbf{\mu}(H)W_i^Q, HW_i^K, HW_i^V) 
\end{align}
\label{formula: vanillaMultihead}
\end{subequations}

\subsubsection{Classifier}\label{subsec: classifier}

Following dimension reduction, the resulting pooled vector ($a$), which is 256-dimensional, is passed to a classification layer. The classifier head is a fully connected layer with randomly initialized weights that maps the vector to a lower-dimensional output based on the number of emotions in the dataset. For example, in IEMOCAP, four emotion classes are used: anger, happiness, sadness, and neutral. Using the Softmax activation function, the most likely emotion is predicted.

\begin{equation}
\text{Emotion} = \text{Softmax}(\text{FC}(a)), \quad \text{Emotion} \in \mathbb{R}^4
\end{equation}

\section{Experiment and results}\label{sec: result and exp}

\subsection{Datasets}\label{subsec: dataset}

\textbf{ShEMO}, or the Sharif Emotional Speech Database, contains 3 hours and 25 minutes of online radio plays consisting of 3,000 utterances from 87 native Persian speakers. Each utterance is labeled with one of six emotions, including anger, fear, happiness, sadness, surprise, and neutral, where anger and neutral account for 70 percent of the utterances, sadness 15 percent, and happiness and surprise together make up less than 15 percent. Fear, as the smallest class, is disregarded in line with previous work. The utterances were divided into 10 groups, where each group consisted of unique speakers.

\vspace{1\baselineskip}
\noindent \textbf{IEMOCAP}, or the Interactive Emotional Dyadic Motion Capture dataset, is a widely used benchmark in SER. It contains approximately 12 hours of dyadic audiovisual data from 10 actors (five male - female pairs) recorded over five sessions. In line with prior work, we restricted our analysis to improvised utterances annotated with four emotional categories: anger, happiness (excited and happy merged), sadness, and neutral. All other emotion labels were excluded, resulting in a dataset of 2,793 utterances. For evaluation, we employed five-fold cross-validation, where in each fold one session was held out for testing and the remaining four sessions were used for training. This established methodology ensures that our results are directly comparable to other studies using the IEMOCAP dataset.

\subsection{Experimental Setup}\label{subsec: setup}

The models are implemented using the PyTorch framework, with pre-trained Whisper weights imported from Hugging Face. Both IEMOCAP and ShEMO datasets are randomly split into batches of size 16 during both the training and testing phases. Based on the dataset partitioning, we perform five-fold and ten-fold cross-validation for IEMOCAP and ShEMO, respectively.

The hyperparameters are selected through a trial-and-error approach. All models are trained for 30 epochs. For ShEMO, both pooling methods use 6 heads, each with a hidden size of 4. For IEMOCAP, the number of heads is set to 12, with each head having a hidden size of 4. The AdamW optimizer is employed to optimize the trainable weights, as it demonstrated strong performance in \cite{nasersharif}. A cosine scheduler is used to adjust the learning rate, which peaks at $10\mathrm{e}{-4}$ with 10\% warmup steps.

\subsection{Results on Pooling Methods}\label{subsec: result}

\begin{table*}[t]
\centering
\caption{Performance comparison of Multi-head Attentive Average Pooling (Attentive) and Multi-head QKV Pooling (QKV). Mean Average Pooling (Mean) serves as the baseline model. Whisper Tiny and Whisper Small encoders were evaluated on the ShEMO and IEMOCAP datasets for all three pooling methods. \textbf{WA} is for Weighted Accuracy and \textbf{UA} is for Unweighted Accuracy.}
\label{tab:pooling_results}
\begin{tabularx}{\linewidth}{l l *{4}{Y}}
\toprule
\textbf{Size} & \textbf{Pooling} & \multicolumn{2}{c}{\textbf{ShEMO}} & \multicolumn{2}{c}{\textbf{IEMOCAP}} \\
\cmidrule(lr){3-4} \cmidrule(lr){5-6}
 & & WA & UA & WA & UA \\
\midrule
\multirow{3}{*}{Tiny}  
 & Mean      & 84.12 $\pm$ 3.60 & 73.66 $\pm$ 3.87 & 68.22 $\pm$ 2.09 & 68.53 $\pm$ 2.35 \\
 & Attentive & 84.71 $\pm$ 4.59 & 74.71 $\pm$ 4.74 & 68.55 $\pm$ 1.79 & 68.67 $\pm$ 1.89 \\
 & QKV       & \textbf{84.81} $\pm$ 3.64 & \textbf{75.14} $\pm$ 4.24 & \textbf{69.37} $\pm$ 1.84 & \textbf{69.38} $\pm$ 1.98 \\
\midrule
\multirow{3}{*}{Small} 
 & Mean      & 88.81 $\pm$ 3.26 & 82.41 $\pm$ 4.02 & 70.52 $\pm$ 1.73 & 70.61 $\pm$ 2.39 \\
 & Attentive & 88.94 $\pm$ 2.11 & 82.86 $\pm$ 4.88 & \textbf{71.98} $\pm$ 1.75 & 72.64 $\pm$ 2.22 \\
 & QKV       & \textbf{89.19} $\pm$ 2.65 & \textbf{83.07} $\pm$ 4.99 & 71.82 $\pm$ 0.22 & \textbf{72.96} $\pm$ 2.22 \\
\bottomrule
\end{tabularx}
\end{table*}

To ensure comparability with previous research in the speech emotion recognition domain, two accuracy metrics, namely Weighted Accuracy and Unweighted Accuracy, are selected for evaluation.

Unweighted Accuracy is defined as the average recall achieved for each class, providing a more balanced performance measure, particularly for datasets such as ShEMO, where testing is conducted on an unbalanced set.

The proposed classification framework is trained and tested using two different Whisper model sizes as the representation extractor. Specifically, we utilized Whisper Tiny and Whisper Small to compare the classification results.

In Section \ref{subsec:drm}, we proposed two attention-based pooling methods. Table \ref{tab:pooling_results} presents their results in comparison to Global Mean Pooling, which serves as the baseline.

Predictably, the attention-based pooling methods outperform simple averaging. This is due to the fact that emotion features are not distributed evenly in an audio sample, with some frames carrying more salient emotional cues than others. Methods that leverage attention address this variability in feature extraction, unlike traditional pooling methods.

QKV pooling achieves the highest accuracy on both ShEMO and IEMOCAP datasets and consistently maintains superior results across different model sizes. However, the lower layers of the QKV-based model exhibit reduced accuracy, indicating that its performance gains are primarily derived from higher-level representations (Section \ref{subsec:layers}).

\subsection{Model Size Factor} \label{subsec:size}

As shown in Table \ref{tab:pooling_results}, the results indicate that the framework achieves superior classification performance when using representations extracted from the larger Whisper model. This was predictable because the Small model also produces a lower word error rate than the Tiny model in Whisper's main task, ASR, thus it generates better representations from the input speech.

\begin{figure}[h!]
	\centering
	\includegraphics[width=0.5\textwidth]{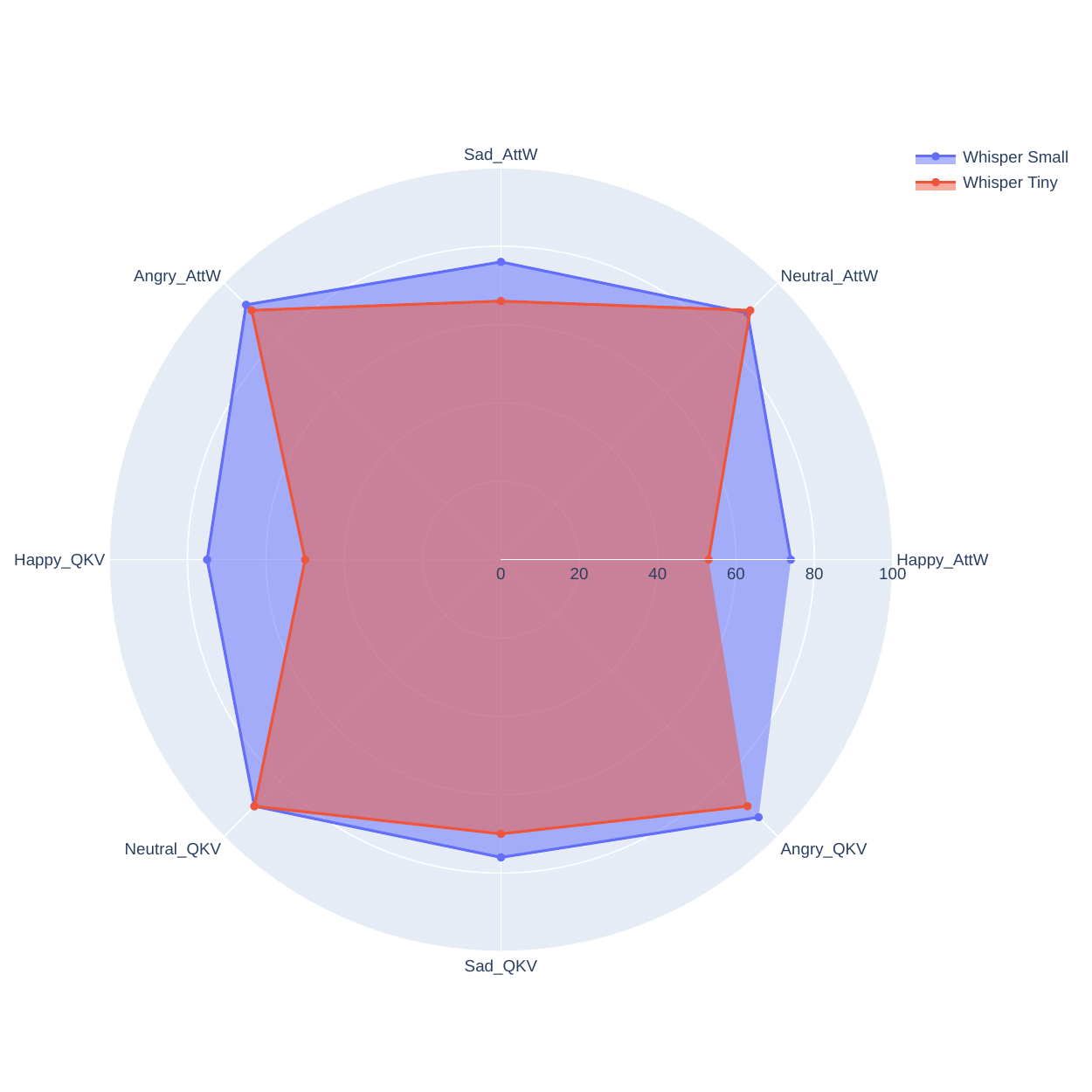}
	\caption{Performance comparison of Whisper Tiny and Whisper Small encoders for Multi-head Attentive Average Pooling (AttW) and Multi-head QKV Pooling (QKV) methods on ShEMO. The radar chart shows that Whisper Small consistently outperforms Whisper Tiny, with particularly notable improvements in categories with fewer samples}
	\label{fig:radar}
\end{figure}

Performance comparison of Whisper Tiny and Whisper Small encoders for Multi-head Attentive Average Pooling (AttW) and Multi-head QKV Pooling (QKV) methods on ShEMO. The radar chart shows that Whisper Small consistently outperforms Whisper Tiny, with particularly notable improvements in categories with fewer samples.

\begin{figure}[h!]
    \centering
    \begin{subfigure}[t]{0.45\textwidth}
        \includegraphics[width=\textwidth]{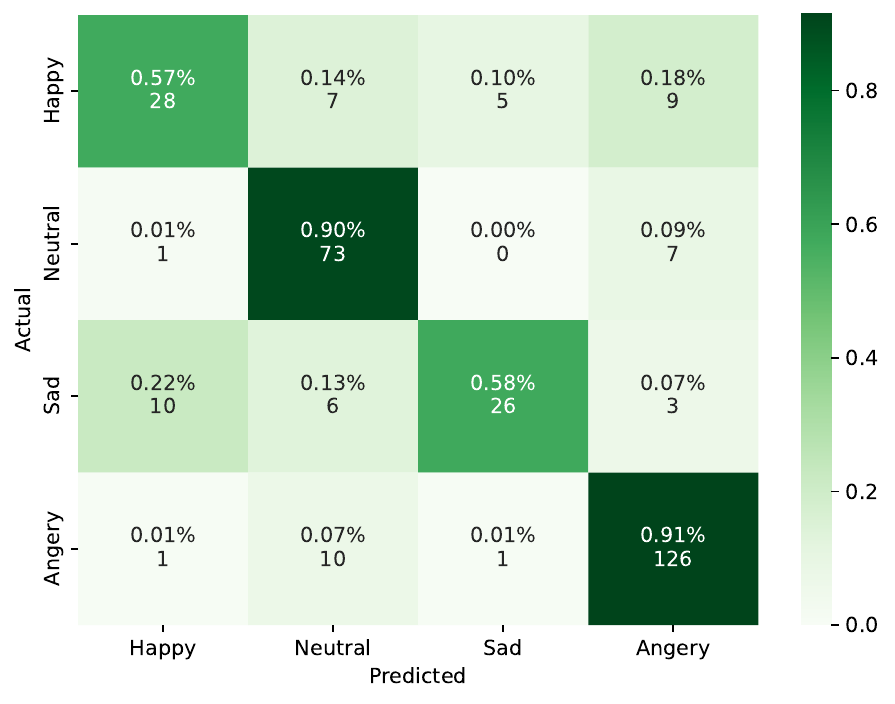}
        \caption{Whisper Tiny}
        \label{fig:subfig1 matrix}
    \end{subfigure}
   \begin{subfigure}[t]{0.45\textwidth}
        \includegraphics[width=\textwidth]{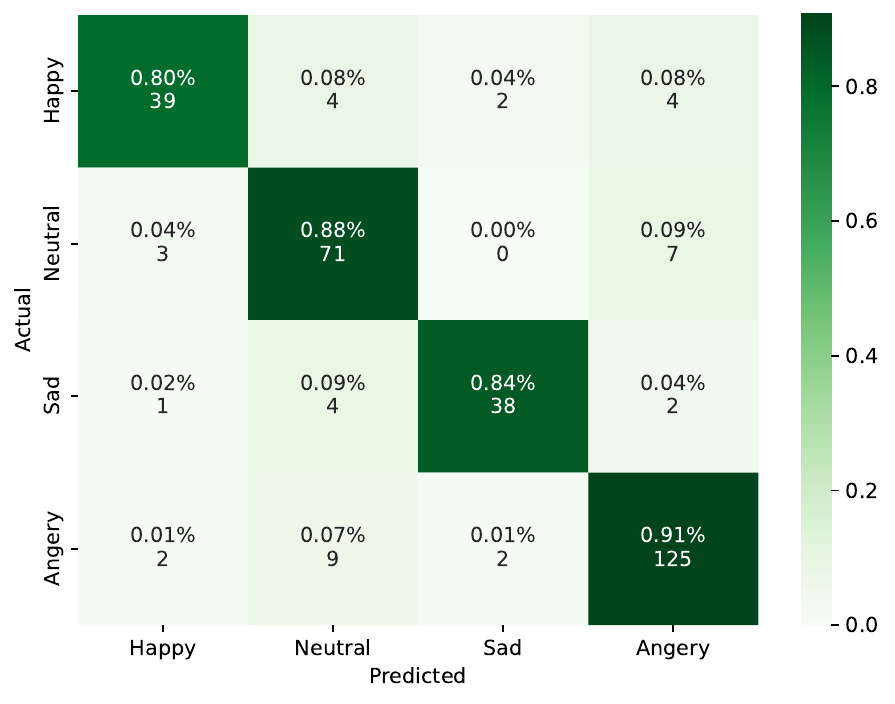}
        \caption{Whisper Small}
        \label{fig:subfig2 matrix}
    \end{subfigure}
    
	\caption{Confusion matrices of classifying ShEMO using different sizes of Whisper for representation extraction.}
    \label{fig:conf_matrices}
\end{figure}

The significant improvement in Unweighted Accuracy for ShEMO, which is a more unbalanced dataset compared to IEMOCAP, suggests that the improved performance of the larger Whisper model can be attributed to its enhanced ability to classify the less frequent categories more effectively. As illustrated in Figure \ref{fig:radar}, the accuracy improvements for the happy and sad categories are more pronounced compared to those for neutral and angry. This trend is also evident when examining the confusion matrices in Figure \ref{fig:conf_matrices}.

\subsection{Cost-accuracy Trade-off}

\begin{table}[t]
\centering
\caption{Comparison of our main results with prior work on ShEMO and IEMOCAP datasets. The best results are highlighted in \textbf{bold}.}
\label{tab:baseline_results}
\begin{tabularx}{\linewidth}{l l Y Y}
\toprule
\textbf{Dataset} & \textbf{Base Model} & \textbf{WA} & \textbf{UA} \\
\midrule
\multirow{6}{*}{ShEMO} 
 & Wav2vec 2.0 Large~\cite{nasersharif}   & 86.80 & 80.60 \\
 & HuBERT Large~\cite{emobox}             & 83.35 & 64.29 \\
 & WavLM Large~\cite{emobox}              & 87.13 & 71.72 \\
 & Data2vec 2.0 Large~\cite{emobox}       & 82.68 & 64.09 \\
 & Whisper Large V3~\cite{emobox}         & \textbf{89.55} & 80.23 \\
 & \textbf{Whisper Small (Ours)}          & 89.19 & \textbf{83.07} \\
\midrule
\multirow{7}{*}{IEMOCAP} 
 & Whisper Large V3~\cite{emobox}         & 72.86 & 73.54 \\
 & HuBERT Large~\cite{emobox}             & 63.10 & 63.87 \\
 & WavLM Large~\cite{emobox}              & 69.07 & 69.47 \\
 & Data2vec 2.0 Large~\cite{emobox}       & 56.23 & 57.30 \\
 & HuBERT X-Large~\cite{dstm}             & 70.65 & 71.46 \\
 & HuBERT X-Large~\cite{mfhca}            & \textbf{74.24} & \textbf{74.57} \\
 & \textbf{Whisper Small (Ours)}          & 71.98 & 72.96 \\
\bottomrule
\end{tabularx}
\end{table}

Obtained results on both datasets are comparable with the latest state-of-the-art (SOTA) research, as shown in Table \ref{tab:baseline_results}. To the best of our knowledge, the highest results obtained on ShEMO are reported in EmoBox \cite{emobox}, with Nasersharif et al. \cite{nasersharif} reporting a slightly higher Unweighted Accuracy.
Our proposed multihead QKV pooling method, however, achieves the best Unweighted Accuracy using a smaller model, highlighting its effectiveness in balancing performance and efficiency.

Although our results do not reach SOTA performance on IEMOCAP compared to other work, we argue that our model requires substantially less computational cost. The lower accuracy compared to the SOTA models can be justified by the significantly higher efficiency and lighter architecture of our approach. It must be noted that the HuBERT X-Large model used in prior studies to extract representations contains approximately ten times more parameters than Whisper Small. Consequently, it requires significantly more computational resources in both training and inference.

Overall, our findings indicate that efficient pooling mechanisms such as QKV pooling can effectively close the gap between smaller and larger pre-trained models. However, as we compared Whisper Tiny and Whisper Small in the previous section, we illustrated the effect of model size on emotion detection. Therefore, we encourage adopting attention-based pooling methods on larger models to achieve further performance improvements.

\subsection{Different Whisper Encoder Layers for Representations} \label{subsec:layers}

\begin{figure*}[h!]
    \centering
    \begin{subfigure}[t]{0.45\textwidth}
        \includegraphics[width=\textwidth]{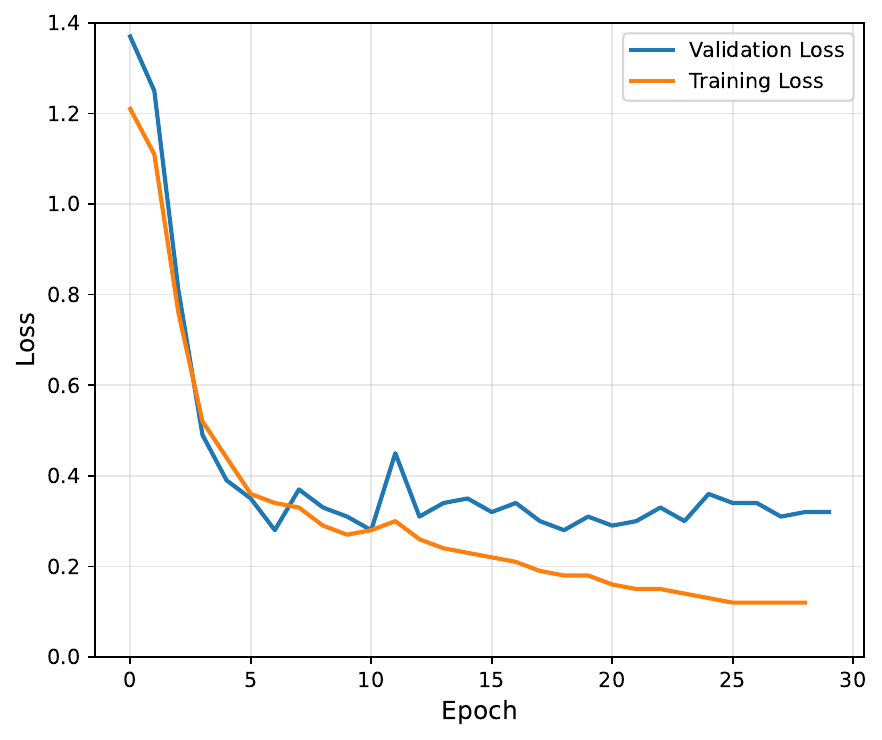}
        \caption{Loss curve for layer 8}
        \label{fig:subfig1}
    \end{subfigure}
    \begin{subfigure}[t]{0.45\textwidth}
        \includegraphics[width=\textwidth]{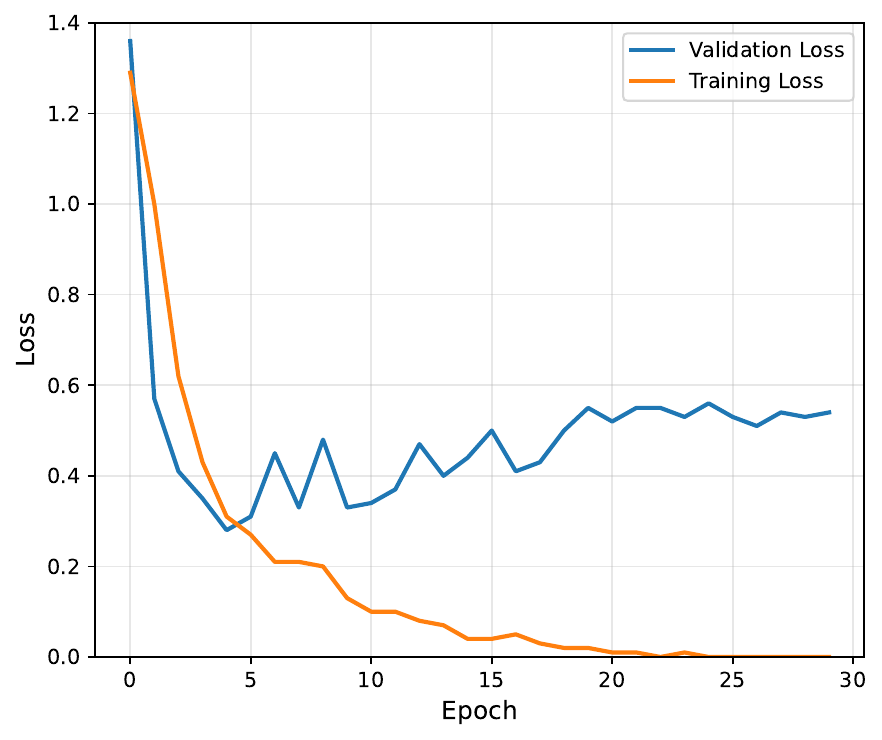}
        \caption{Loss curve for layer 12}
        \label{fig:subfig2}
    \end{subfigure}
    \begin{subfigure}[t]{0.45\textwidth}
        \includegraphics[width=\textwidth]{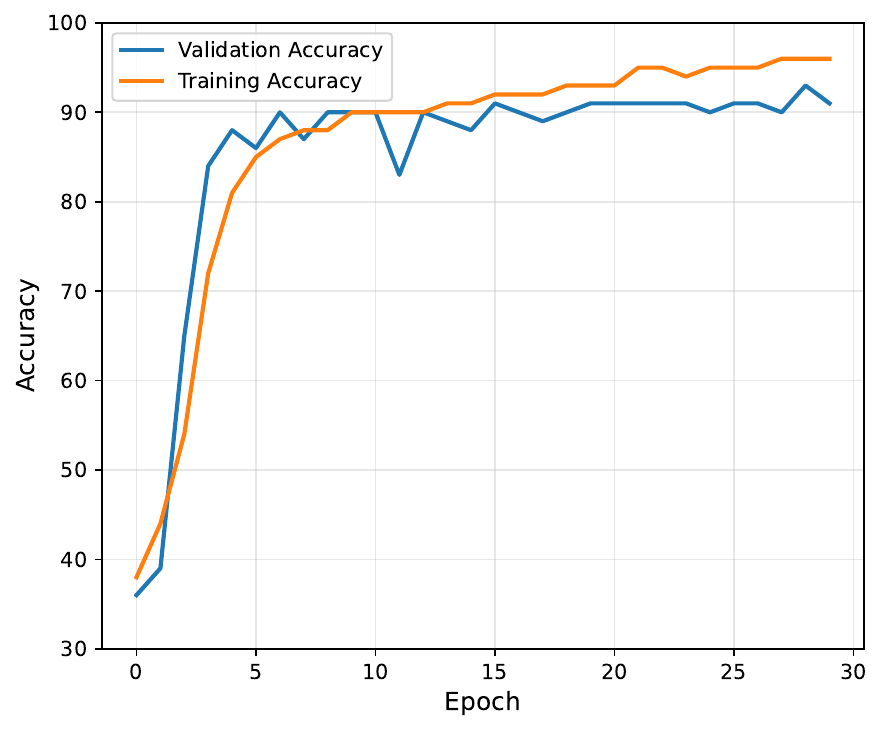}
        \caption{Accuracy curve for layer 8}
        \label{fig:subfig2}
    \end{subfigure}
    \begin{subfigure}[t]{0.45\textwidth}
        \includegraphics[width=\textwidth]{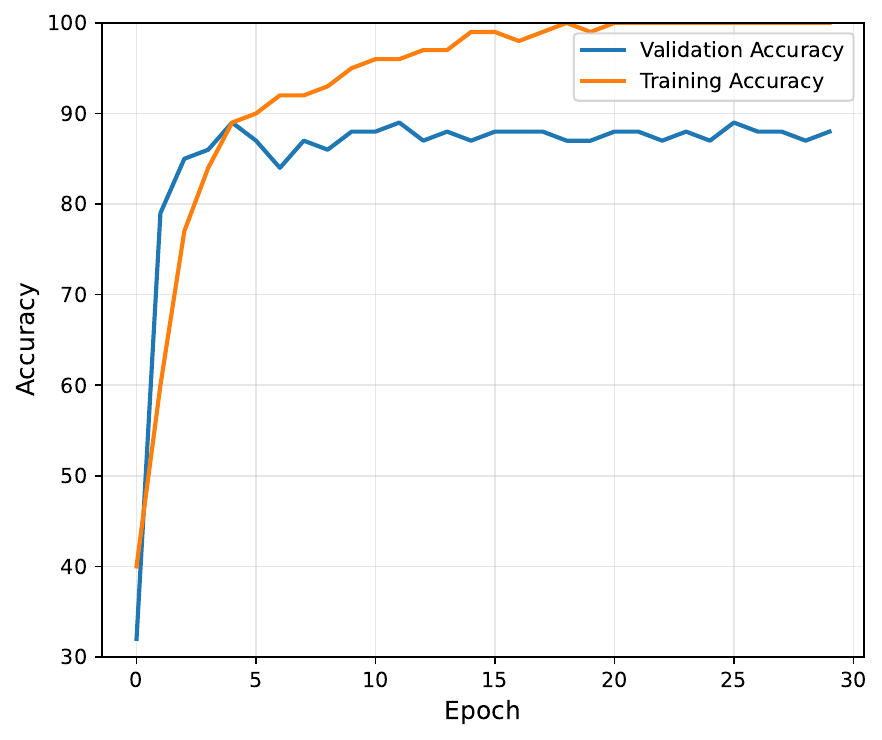}
        \caption{Accuracy curve for layer 12}
        \label{fig:subfig2}
    \end{subfigure}
    \caption{Comparing learning speed using different layers of whisper encoder as representation in ShEMO}
    \label{fig:loss_and_accuracy}
\end{figure*}

According to Pepino et al. \cite{pepino}, different layers of pre-trained models capture different types of information during training. While the last encoder layer is generally best for producing representations for the main pre-training task, the middle layers often contain more useful information for tasks such as SER.

Whisper, the pre-trained model used in this study, was trained in a multitask setting to perform tasks including ASR, voice activity detection (VAD), language detection, and speech translation. It was not explicitly trained for emotion detection in speech. Moreover, in the multilingual model, different encoder layers may have acquired different types of information depending on the input language.

Previous work has applied Whisper for speech classification and SER \cite{emobox, abuse_detection}, but the experiments were limited to the model representations of the final layer. To the best of our knowledge, this is the first work to investigate the model representations of each layer in depth. We study and report the performance of all four layers of Whisper Tiny and all twelve layers of Whisper Small.

Models utilizing the final layer of Whisper reached their maximum learning potential much faster but also tended to overfit earlier. In contrast, models based on intermediate and lower layers required more epochs to achieve their full potential. This observation, shown in Figure \ref{fig:loss_and_accuracy}, held consistently across both the IEMOCAP and ShEMO datasets.

\begin{figure}[h!]
    \centering
    \begin{subfigure}[t]{0.45\textwidth}
        \includegraphics[width=\textwidth]{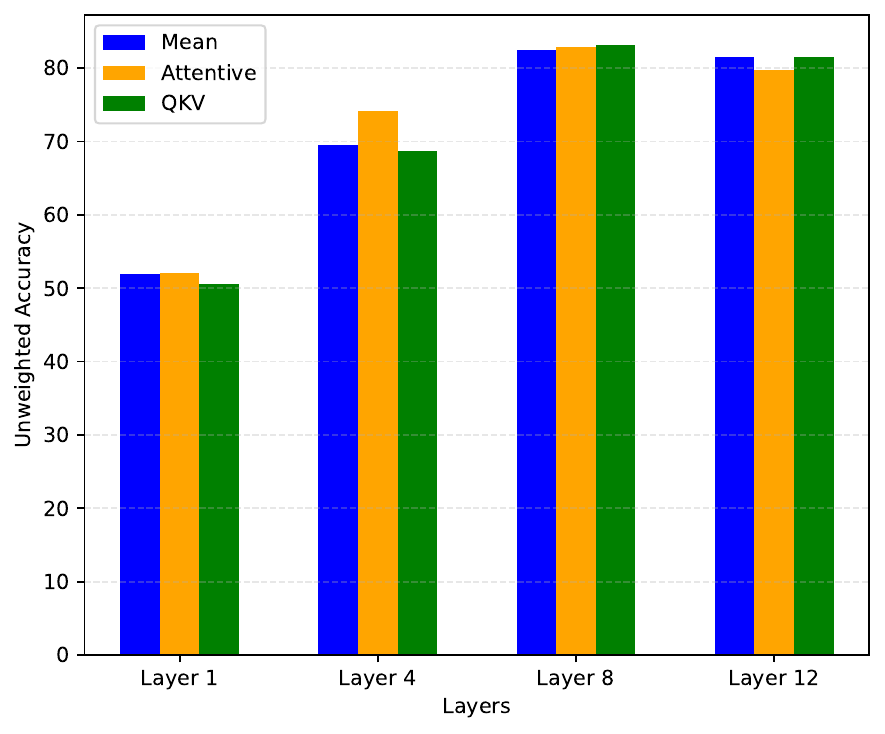}
        \caption{ShEMO}
        \label{fig:subfig1 matrix}
    \end{subfigure}
   \begin{subfigure}[t]{0.45\textwidth}
        \includegraphics[width=\textwidth]{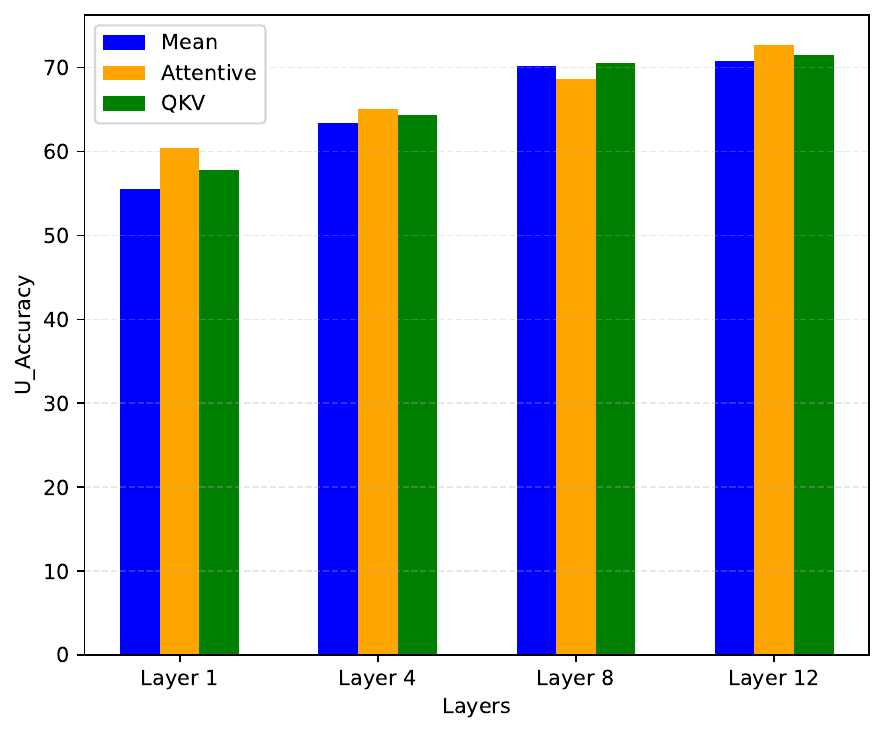}
        \caption{IEMOCAP}
        \label{fig:subfig2 matrix}
    \end{subfigure}
    
    \caption{Performance comparison of Mean Average Pooling (Mean), Multi-head Attentive Average Pooling (Attentive) and Multi-head QKV Pooling (QKV) when using 4 different layers of Whisper Small Encoders.}
    \label{fig:layers_compare}
\end{figure}

According to Figure \ref{fig:layers_compare}, for the IEMOCAP dataset, in most cases, using the output of the last Whisper encoder layer yields the best performance. However, for the ShEMO dataset, using the middle layers as speech representations yields better results. Complete experimental results are presented in Appendix \ref{secA1}.

\section{Discussion and Future Work}\label{sec: Discussion}

In the projection stage, a substantial dimensionality reduction occurs. For instance, in the small model, 768-dimensional vectors are reduced to 256 dimensions through a fully connected layer with randomly initialized weights. Later, in the classification stage, the 256-dimensional vector is mapped to a 4-dimensional output, again through a fully connected layer with randomly initialized weights. This introduces a large number of trainable parameters, for example, in the small model 197,632 weights are solely dedicated to dimensionality reduction. Such parameters not only require random initialization and training but also demand larger datasets than those available in our experiments, while increasing model overhead. Therefore, employing more creative and efficient methods for dimensionality reduction could potentially lead to higher accuracy and better model optimization.

One promising direction is the integration of an ASR system to generate transcriptions from speech, using text as an additional modality for emotion recognition. Given that the Whisper decoder inherently maps speech representations to text and benefits from pretrained weights, it may eliminate the need for an additional embedding model such as BERT. In other words, the decoder produces intermediate vector representations that can directly be exploited for emotion classification, avoiding the computationally expensive two-step process of first generating text and then embedding it again with a separate language model.

Another line of exploration is multimodal Emotion Recognition. For example, datasets such as IEMOCAP or MELD \cite{meld_dataset} contain not only speech but also visual data. Leveraging facial gestures as an auxiliary modality could improve recognition performance.

In this study, we focused solely on using the output of individual Whisper encoder layers as standalone speech representations and compared their performance. Since some layers exhibited only marginal differences in performance, it suggests that each layer encodes useful but slightly distinct information for emotion classification. Consequently, designing an attention-based mechanism that aggregates information across all encoder layers into a unified representation matrix may yield better results. However, this would increase model complexity and require more powerful hardware (e.g., graphical processing units with higher VRAM).

Compared with competing approaches that rely on models such as HuBERT X-Large (1B parameters) or Wav2Vec 2.0 (300M parameters) for SER on datasets like IEMOCAP or ShEMO, the Whisper-small model with an encoder of only 88M parameters represents a lightweight yet effective alternative. Even when performance is comparable, the use of Whisper-small offers a more efficient and practical solution.

Furthermore, our findings on ShEMO suggest that intermediate encoder layers of Whisper yield better performance than the final layers. This indicates that intermediate-layer representations can serve as effective speech embeddings while reducing computational cost, as the final layers need not be activated. This further enhances the efficiency and practicality of Whisper-based SER systems.

\section{Conclusion}\label{sec: conclusion}

In this work, we explore the research gap for achieving a pooling method that prevents the loss of valuable information from a speech representation. We chose Whisper for its state-of-the-art performance and multilingual abilities to generate representations from audio samples. The multilingual abilities of Whisper also allow us to contribute to SER in Persian, as a low-resource language, as well as English, enabling us to compare our results globally and with more studies. We use ShEMO and IEMOCAP for Persian and English, respectively. After attaining the audio representations using Whisper encoders, we applied a pooling method based on an attention mechanism. The first method, Multi-head Attentive Average Pooling, extends existing attentive statistics pooling by integrating a multi-head attention mechanism to assign importance weights to different frames within an utterance, considering that some frames contain more informative features for emotion prediction than others. The second method, Multi-head QKV Attention Pooling, utilizes a multi-head QKV (Query, Key, Value) attention mechanism for pooling, but since the classification task lacks prior generated tokens usually required for the Query (Q), the approach conditions the Query on the globally average-pooled representation of the audio sample. Our experiments show that QKV achieves SOTA results on the Shemo dataset for unweighted accuracy, which is significant in an unbalanced dataset such as Shemo. Also, compared to the SOTA for IEMOCAP, our method consumes less computational cost and appears to be a good trade-off between accuracy and cost.\\ 
\pagebreak

\newcounter{savetable}
\setcounter{savetable}{\value{table}}  

\bibliography{sn-bibliography}
\pagebreak
\begin{appendices}
\setcounter{table}{\value{savetable}}  
\section{Results on Whisper Encoder Layers}\label{secA1}
\counterwithout{table}{section} 
In Tables \ref{tab:full_pooling_results_tiny} and  \ref{tab:full_pooling_results_small}, we have included the Weighted and Unweighted Accuracy of each transformer layer in the Whisper model. We experimented with Whisper Tiny and Whisper Large, with 4 and 12 transformer blocks, respectively.

\begin{table*}[h!]
\centering
\caption{Performance comparison of different pooling methods and layers for Whisper Tiny model configurations on IEMOCAP and ShEMO datasets.}
\label{tab:full_pooling_results_tiny}
\small 
\setlength{\tabcolsep}{3pt} 
\begin{tabularx}{\linewidth}{c *{7}{Y}}
\toprule
 & \multicolumn{3}{c}{\textbf{IEMOCAP}} & \multicolumn{3}{c}{\textbf{ShEMO}} \\
\cmidrule(lr){2-4} \cmidrule(lr){5-7}
\textbf{Layer} & \textbf{WA} & \textbf{UA} & \textbf{F1} & \textbf{WA} & \textbf{UA} & \textbf{F1} \\
\midrule
\multicolumn{7}{c}{\textbf{Whisper Tiny + Mean Pooling}} \\
\midrule
1 & 54.62 $\pm$ 2.27 & 55.40 $\pm$ 2.37 & 54.06 $\pm$ 2.23 & 70.94 $\pm$ 5.25 & 50.47 $\pm$ 2.87 & 67.08 $\pm$ 6.15 \\
2 & 60.85 $\pm$ 1.30 & 61.76 $\pm$ 1.54 & 60.08 $\pm$ 1.65 & 80.88 $\pm$ 4.20 & 66.94 $\pm$ 4.82 & 79.54 $\pm$ 4.72 \\
3 & 66.44 $\pm$ 1.93 & 66.92 $\pm$ 2.62 & 66.11 $\pm$ 2.07 & 83.26 $\pm$ 4.00 & 71.74 $\pm$ 3.78 & 82.48 $\pm$ 4.27 \\
4 & 68.22 $\pm$ 2.09 & 68.53 $\pm$ 2.35 & 67.98 $\pm$ 2.13 & 84.12 $\pm$ 3.60 & 73.66 $\pm$ 3.87 & 83.47 $\pm$ 3.77 \\
\midrule
\multicolumn{7}{c}{\textbf{Whisper Tiny + Attentive Average Pooling}} \\
\midrule
1 & 58.42 $\pm$ 2.46 & 58.88 $\pm$ 2.80 & 57.72 $\pm$ 2.89 & 78.35 $\pm$ 4.13 & 60.22 $\pm$ 6.33 & 76.42 $\pm$ 4.46 \\
2 & 64.83 $\pm$ 1.84 & 64.84 $\pm$ 2.83 & 64.25 $\pm$ 2.11 & 80.33 $\pm$ 3.72 & 64.73 $\pm$ 6.21 & 78.98 $\pm$ 4.10 \\
3 & 66.67 $\pm$ 1.44 & 67.47 $\pm$ 1.86 & 66.42 $\pm$ 1.66 & 84.71 $\pm$ 4.59 & 74.71 $\pm$ 4.74 & 84.29 $\pm$ 4.53 \\
4 & 68.57 $\pm$ 2.17 & 68.69 $\pm$ 2.18 & 68.50 $\pm$ 2.42 & 83.09 $\pm$ 3.56 & 73.51 $\pm$ 3.48 & 82.78 $\pm$ 3.65 \\
\midrule
\multicolumn{7}{c}{\textbf{Whisper Tiny + QKV Pooling}} \\
\midrule
1 & 54.74 $\pm$ 2.22 & 56.43 $\pm$ 0.92 & 53.45 $\pm$ 2.65 & 73.95 $\pm$ 5.65 & 52.78 $\pm$ 4.72 & 70.24 $\pm$ 6.25 \\
2 & 61.80 $\pm$ 1.60 & 62.51 $\pm$ 2.56 & 61.22 $\pm$ 1.97 & 79.89 $\pm$ 4.39 & 63.67 $\pm$ 6.01 & 78.06 $\pm$ 5.19 \\
3 & 66.94 $\pm$ 1.41 & 67.88 $\pm$ 1.77 & 66.58 $\pm$ 1.79 & 84.81 $\pm$ 3.64 & 74.90 $\pm$ 4.48 & 84.26 $\pm$ 4.06 \\
4 & 69.37 $\pm$ 1.84 & 69.38 $\pm$ 1.98 & 69.31 $\pm$ 1.91 & 83.32 $\pm$ 3.86 & 75.14 $\pm$ 4.24 & 82.97 $\pm$ 3.88 \\
\bottomrule
\end{tabularx}
\end{table*}

\begin{table*}[h!]
\centering
\caption{Performance comparison of different pooling methods and layers for Whisper Small model configurations on IEMOCAP and ShEMO datasets.}
\label{tab:full_pooling_results_small}
\small 
\setlength{\tabcolsep}{3pt} 
\begin{tabularx}{\linewidth}{c *{7}{Y}}
\toprule
 & \multicolumn{3}{c}{\textbf{IEMOCAP}} & \multicolumn{3}{c}{\textbf{ShEMO}} \\
\cmidrule(lr){2-4} \cmidrule(lr){5-7}
\textbf{Layer} & \textbf{WA} & \textbf{UA} & \textbf{F1} & \textbf{WA} & \textbf{UA} & \textbf{F1} \\
\midrule
\multicolumn{7}{c}{\textbf{Whisper Small + Mean Pooling}} \\
\midrule
1 & 55.58 $\pm$ 1.30 & 55.54 $\pm$ 2.06 & 55.29 $\pm$ 1.37 & 73.20 $\pm$ 6.78 & 51.97 $\pm$ 6.86 & 69.41 $\pm$ 7.98 \\
2 & 58.42 $\pm$ 1.64 & 59.90 $\pm$ 2.58 & 57.79 $\pm$ 1.86 & 77.15 $\pm$ 5.99 & 58.08 $\pm$ 4.33 & 74.57 $\pm$ 6.66 \\
3 & 59.88 $\pm$ 1.64 & 59.94 $\pm$ 1.65 & 59.27 $\pm$ 2.21 & 79.54 $\pm$ 5.51 & 62.41 $\pm$ 7.58 & 77.35 $\pm$ 6.51 \\
4 & 62.43 $\pm$ 1.81 & 63.43 $\pm$ 1.76 & 61.93 $\pm$ 2.27 & 82.53 $\pm$ 4.55 & 69.48 $\pm$ 4.44 & 81.43 $\pm$ 5.13 \\
5 & 66.68 $\pm$ 1.26 & 67.48 $\pm$ 2.49 & 66.42 $\pm$ 1.18 & 85.20 $\pm$ 3.96 & 76.03 $\pm$ 4.16 & 84.71 $\pm$ 4.23 \\
6 & 67.77 $\pm$ 1.65 & 67.42 $\pm$ 2.52 & 67.78 $\pm$ 1.69 & 85.89 $\pm$ 3.37 & 77.30 $\pm$ 3.32 & 85.65 $\pm$ 3.44 \\
7 & 68.27 $\pm$ 1.48 & 67.95 $\pm$ 1.83 & 67.99 $\pm$ 1.70 & 88.19 $\pm$ 2.46 & 79.84 $\pm$ 4.58 & 87.84 $\pm$ 2.53 \\
8 & 69.99 $\pm$ 1.26 & 70.24 $\pm$ 1.84 & 69.88 $\pm$ 1.26 & 88.81 $\pm$ 3.26 & 82.41 $\pm$ 4.02 & 88.59 $\pm$ 3.25 \\
9 & 70.05 $\pm$ 1.79 & 70.16 $\pm$ 2.37 & 69.94 $\pm$ 1.95 & 87.41 $\pm$ 3.76 & 78.40 $\pm$ 7.80 & 87.15 $\pm$ 3.86 \\
10 & 70.52 $\pm$ 1.73 & 70.61 $\pm$ 2.39 & 70.43 $\pm$ 1.76 & 87.07 $\pm$ 3.52 & 77.92 $\pm$ 7.15 & 86.64 $\pm$ 3.73 \\
11 & 70.65 $\pm$ 1.35 & 70.66 $\pm$ 2.35 & 70.56 $\pm$ 1.45 & 86.65 $\pm$ 3.35 & 77.55 $\pm$ 5.86 & 86.22 $\pm$ 3.47 \\
12 & 70.69 $\pm$ 1.58 & 70.78 $\pm$ 2.37 & 70.62 $\pm$ 1.05 & 87.61 $\pm$ 3.15 & 81.65 $\pm$ 5.71 & 87.56 $\pm$ 3.09 \\
\midrule
\multicolumn{7}{c}{\textbf{Whisper Small + Attentive Average Pooling}} \\
\midrule
1 & 59.07 $\pm$ 2.73 & 60.47 $\pm$ 1.83 & 58.13 $\pm$ 3.44 & 73.15 $\pm$ 6.61 & 52.13 $\pm$ 5.49 & 69.62 $\pm$ 7.42 \\
2 & 61.75 $\pm$ 1.49 & 62.54 $\pm$ 2.11 & 61.36 $\pm$ 1.64 & 77.55 $\pm$ 5.87 & 58.66 $\pm$ 5.28 & 74.71 $\pm$ 6.70 \\
3 & 63.23 $\pm$ 1.51 & 64.32 $\pm$ 1.52 & 62.48 $\pm$ 2.26 & 80.69 $\pm$ 4.96 & 65.48 $\pm$ 4.18 & 78.87 $\pm$ 5.84 \\
4 & 64.96 $\pm$ 1.48 & 65.03 $\pm$ 2.89 & 64.51 $\pm$ 1.57 & 83.18 $\pm$ 4.77 & 74.19 $\pm$ 3.58 & 82.80 $\pm$ 4.80 \\
5 & 66.83 $\pm$ 0.87 & 67.70 $\pm$ 1.44 & 66.61 $\pm$ 0.82 & 85.19 $\pm$ 3.40 & 76.50 $\pm$ 4.61 & 84.84 $\pm$ 3.54 \\
6 & 67.51 $\pm$ 2.22 & 67.46 $\pm$ 2.06 & 67.53 $\pm$ 2.53 & 85.98 $\pm$ 2.67 & 78.02 $\pm$ 5.47 & 85.75 $\pm$ 2.86 \\
7 & 69.36 $\pm$ 2.09 & 70.81 $\pm$ 2.63 & 69.28 $\pm$ 2.14 & 88.31 $\pm$ 2.74 & 81.89 $\pm$ 3.18 & 88.12 $\pm$ 2.78 \\
8 & 68.79 $\pm$ 1.97 & 68.70 $\pm$ 2.25 & 68.66 $\pm$ 1.94 & 88.94 $\pm$ 2.11 & 82.86 $\pm$ 4.88 & 88.79 $\pm$ 2.18 \\
9 & 68.29 $\pm$ 2.26 & 68.79 $\pm$ 2.49 & 68.05 $\pm$ 2.23 & 88.63 $\pm$ 3.16 & 82.55 $\pm$ 5.21 & 88.44 $\pm$ 3.20 \\
10 & 69.35 $\pm$ 2.43 & 69.68 $\pm$ 2.84 & 69.21 $\pm$ 2.45 & 87.02 $\pm$ 2.96 & 79.17 $\pm$ 5.44 & 86.59 $\pm$ 3.37 \\
11 & 71.50 $\pm$ 3.09 & 71.11 $\pm$ 3.09 & 71.47 $\pm$ 3.14 & 88.19 $\pm$ 2.66 & 80.22 $\pm$ 6.32 & 88.02 $\pm$ 2.73 \\
12 & 71.98 $\pm$ 1.75 & 72.64 $\pm$ 2.22 & 71.89 $\pm$ 1.64 & 86.71 $\pm$ 2.71 & 79.75 $\pm$ 3.60 & 86.44 $\pm$ 2.66 \\
\midrule
\multicolumn{7}{c}{\textbf{Whisper Small + QKV Pooling}} \\
\midrule
1 & 55.85 $\pm$ 1.77 & 57.82 $\pm$ 2.13 & 54.81 $\pm$ 2.05 & 73.06 $\pm$ 6.32 & 50.59 $\pm$ 5.55 & 68.90 $\pm$ 6.77 \\
2 & 58.15 $\pm$ 2.00 & 59.89 $\pm$ 2.89 & 57.61 $\pm$ 2.23 & 76.91 $\pm$ 6.11 & 57.21 $\pm$ 4.31 & 73.91 $\pm$ 7.07 \\
3 & 60.37 $\pm$ 1.75 & 62.17 $\pm$ 2.64 & 59.72 $\pm$ 2.21 & 78.80 $\pm$ 5.89 & 61.67 $\pm$ 5.86 & 76.58 $\pm$ 7.21 \\
4 & 62.86 $\pm$ 1.58 & 64.40 $\pm$ 1.92 & 62.52 $\pm$ 1.89 & 82.43 $\pm$ 5.26 & 68.67 $\pm$ 6.71 & 81.07 $\pm$ 5.98 \\
5 & 66.31 $\pm$ 0.91 & 67.87 $\pm$ 2.19 & 65.84 $\pm$ 1.01 & 85.24 $\pm$ 4.34 & 75.83 $\pm$ 5.98 & 84.95 $\pm$ 4.43 \\
6 & 67.90 $\pm$ 2.29 & 68.20 $\pm$ 2.86 & 67.90 $\pm$ 2.37 & 85.84 $\pm$ 3.80 & 78.17 $\pm$ 4.53 & 85.61 $\pm$ 3.93 \\
7 & 69.46 $\pm$ 1.46 & 69.23 $\pm$ 2.23 & 69.30 $\pm$ 1.47 & 88.16 $\pm$ 2.88 & 81.78 $\pm$ 3.04 & 88.01 $\pm$ 2.83 \\
8 & 70.15 $\pm$ 1.42 & 70.55 $\pm$ 2.06 & 69.95 $\pm$ 1.45 & 89.19 $\pm$ 2.65 & 83.07 $\pm$ 4.99 & 89.04 $\pm$ 2.62 \\
9 & 70.66 $\pm$ 1.34 & 70.70 $\pm$ 1.70 & 70.67 $\pm$ 1.36 & 88.45 $\pm$ 3.47 & 82.75 $\pm$ 4.70 & 88.35 $\pm$ 3.47 \\
10 & 71.83 $\pm$ 1.44 & 72.04 $\pm$ 1.19 & 71.74 $\pm$ 1.45 & 88.55 $\pm$ 3.29 & 82.85 $\pm$ 3.07 & 88.40 $\pm$ 3.27 \\
11 & 71.79 $\pm$ 2.20 & 72.96 $\pm$ 2.22 & 71.50 $\pm$ 2.25 & 88.02 $\pm$ 3.13 & 78.76 $\pm$ 6.90 & 87.63 $\pm$ 3.32 \\
12 & 71.04 $\pm$ 1.61 & 71.51 $\pm$ 1.93 & 70.96 $\pm$ 1.46 & 87.34 $\pm$ 2.43 & 81.44 $\pm$ 3.42 & 87.17 $\pm$ 2.38 \\
\bottomrule
\end{tabularx}
\end{table*}

%



\end{appendices}


\end{document}